\documentclass[nohyperref]{article}

\usepackage{microtype}
\usepackage{graphicx}
\usepackage{subfigure}
\usepackage{booktabs} 

\usepackage{hyperref}

\usepackage[accepted]{icml2022}

\usepackage{amsmath}
\usepackage{amssymb}
\usepackage{mathtools}
\usepackage{amsthm}
\usepackage{bm}
\usepackage{multirow}
\usepackage{multicol}

\usepackage[capitalize,noabbrev]{cleveref}

\theoremstyle{plain}

\theoremstyle{definition}

\theoremstyle{remark}

\newcommand{\etal}{\emph{et al.}}
\newcommand{\eg}{\emph{e.g.}}
\newcommand{\ie}{\emph{i.e.}}

\usepackage[textsize=tiny]{todonotes}

\icmltitlerunning{Efficient Representation Learning via Adaptive Context Pooling}

\begin{document}

\twocolumn[
\icmltitle{Efficient Representation Learning via Adaptive Context Pooling}




\begin{icmlauthorlist}
\icmlauthor{Chen Huang}{apple}
\icmlauthor{Walter Talbott}{apple}
\icmlauthor{Navdeep Jaitly}{apple}
\icmlauthor{Josh Susskind}{apple}
\end{icmlauthorlist}

\icmlaffiliation{apple}{Apple Inc., Cupertino, United States}

\icmlcorrespondingauthor{Chen Huang}{chen-huang@apple.com}

\icmlkeywords{ICML, ContextPool, Efficient Representation Learning}

\vskip 0.3in
]



\printAffiliationsAndNotice{}  

\begin{abstract}
Self-attention mechanisms model long-range context by using pairwise attention between all input tokens. In doing so, they assume a fixed attention granularity defined by the individual tokens (\eg,~text characters or image pixels), which may not be optimal for modeling complex dependencies at higher levels. In this paper, we propose \emph{ContextPool} to address this problem by adapting the attention granularity for each token. Inspired by the success of ConvNets that are combined with pooling to capture long-range dependencies, we learn to pool neighboring features for each token before computing attention in a given attention layer. The pooling weights and support size are adaptively determined, allowing the pooled features to encode meaningful context with varying scale. We show that ContextPool makes attention models more expressive, achieving strong performance often with fewer layers and thus significantly reduced cost. Experiments validate that our ContextPool module, when plugged into transformer models, matches or surpasses state-of-the-art performance using less compute on several language and image benchmarks, outperforms recent works with learned context sizes or sparse attention patterns, and is also applicable to ConvNets for efficient feature learning.


\end{abstract}

\section{Introduction}
Transformers~\cite{NIPS2017_3f5ee243} have achieved great success in the domains of natural language processing (NLP)~\cite{devlin-2019-bert} and computer vision~\cite{dosovitskiy2021an}. These models benefit from the self-attention mechanism, which computes correlations between all pairs of tokens in an input sequence. Self-attention enables transformers to capture long-range context, which is important in both language and vision tasks. 

\begin{figure}[!t]
\vskip 0.2in
\begin{center}
\centerline{\includegraphics[width=1.1\columnwidth]{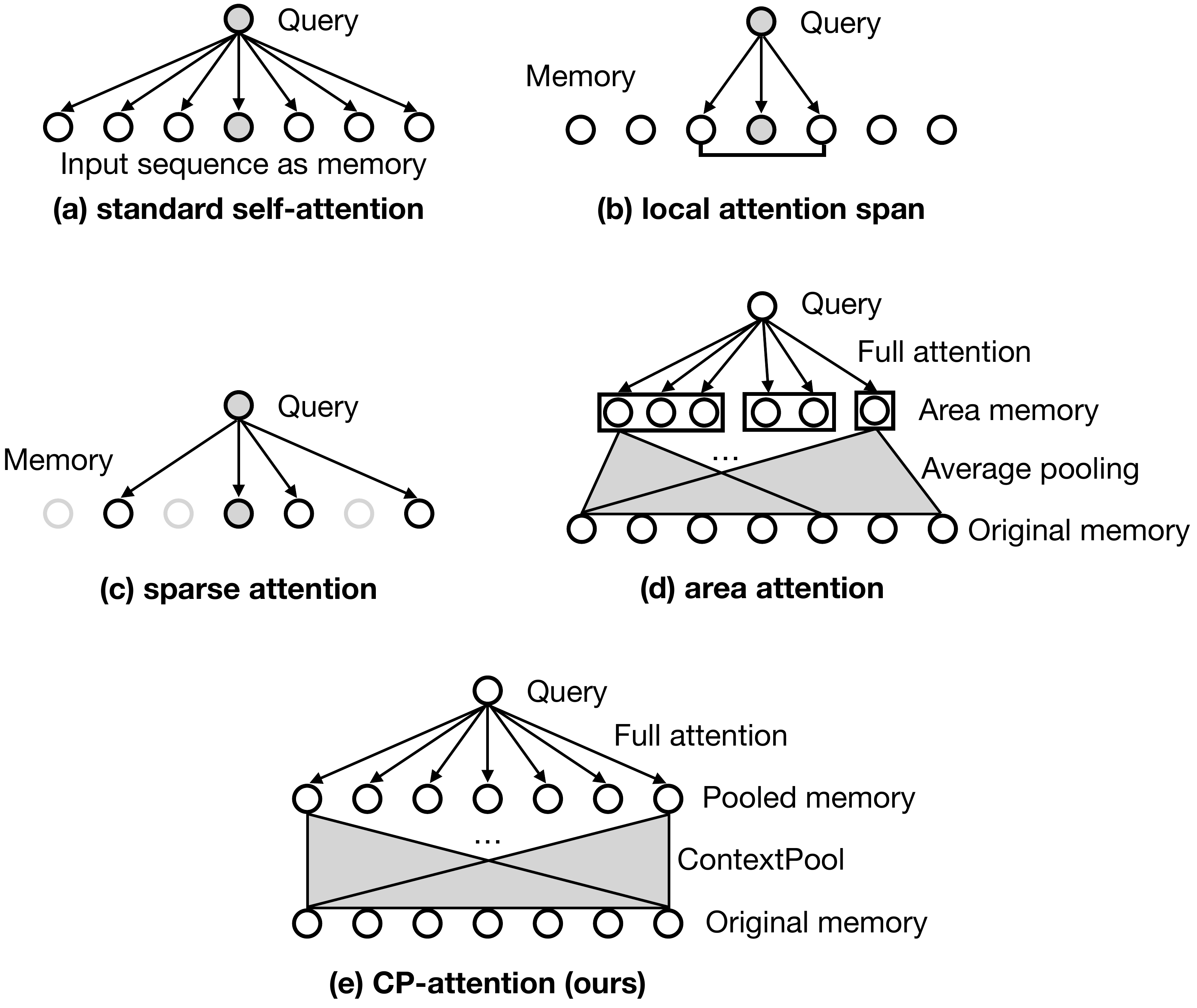}}
\caption{Comparing transformers with (a) standard self-attention~\cite{NIPS2017_3f5ee243}, (b-c) efficient attention mechanisms with localized~\cite{yang-2018} or other sparsity patterns~\cite{LiJXZCWY19} that lose the full-attention capacity, and (d) area attention~\cite{pmlr-v97-li19e} that maintains an extra memory formed by average pooling with a predefined set of pool sizes. (e) Our ContextPool learns to pool with adaptive weighting and support size for each token in-place, before computing full attention.}
\label{fig:compare_scheme}
\end{center}
\vskip -0.3in
\end{figure}

However, each attention layer uses pairwise relationships between individual tokens (\eg,~text characters and image pixels), which implies a fixed granularity for attention. This ignores the context around each token, which can vary substantially in scale in the vision and language domains,~\eg,~from characters to words and from phrases to sentence. Therefore, self-attention with fixed granularity can be fundamentally limited for modeling complex distribution of contextual dependencies, and several layers of self-attention might be needed to make up for this fixed granularity.

Recent vision transformers such as Swin transformer~\cite{liu2021Swin} and PVT~\cite{Wang_2021_ICCV} adopt a hierarchical architecture to compute self-attention at various scales. However, such attention scale or granularity is predetermined rather than learned. Similarly, Li~\etal~\yrcite{pmlr-v97-li19e} proposed to use a predefined set of pooling sizes to form a multi-scale `area memory', which accounts for varying context range but in fixed architecture. In BP-Transformer~\cite{1911-04070}, a fine-to-coarse attention is computed from multi-scale attention spans via automatic binary partitioning, but the resulting local span sequences might still hurt the capacity of \emph{full} attention.

In this paper we propose \emph{ContextPool}, a drop-in and low-cost module for both the transformer and convolutional networks (ConvNets) to enhance their capacity to model long-range context with dynamic scales, and hence to facilitate efficient representation learning. The idea behind ContextPool is in general inspired by ConvNets, which have local receptive fields and pooling operations. 
Here we similarly learn to pool neighboring features for each token at every attention layer before computing full attention in transformers. Importantly, the pooling weights and support size are input-adaptive. This allows the pooled features to encode meaningful context with dynamic scale. As a result, self-attention among pooled features can explicitly capture high-level dependencies between contexts.

We show our simple ContextPool makes attention models more expressive, achieving strong performance often with fewer layers. This leads to significantly reduced cost without much sacrifice in accuracy. On the other hand, when we can maintain the same level of compute cost, ContextPool consistently improves performance as it can model longer range of context. When compared to recent transformer models that reduce cost by sparsifying the attention matrix~\cite{yang-2018,sukhbaatar2019,child2019sparsetransformer,ainslie2020etc}, our ContextPool method preserves the full attention capability (see comparison in Fig.~\ref{fig:compare_scheme}) and can be considered orthogonal to those efficient techniques.

Experiments show that our ContextPool module significantly improves transformer models in terms of performance-cost trade-off, matching or surpassing state-of-the-art performance with less compute on several language and image benchmarks. ContextPool also outperforms recent works with adaptive context size or sparse attention, and is applicable to ConvNets for efficient representation learning. To summarize, our main \textbf{contributions} are:
\begin{itemize}
\item We introduce ContextPool to encode varying-sized context for each token in an attention layer, giving rise to self-attention with adaptive granularity to model high-level dependencies.
\item We show ContextPool-based transformers achieve competitive performance with much less compute on several language and image benchmarks, and outperform prior works with adaptive context size or sparse attention patterns.
\item ContextPool is applicable to ConvNets with strong image recognition performance, showing its promise to be a generic module for efficient representation learning.
\end{itemize}

\section{Related Work}
\textbf{Context in Transformers} is captured by the attention mechanism between all pairs of tokens from the entire input sequence. When the network goes deeper, high-level contextual dependencies emerge. However, full attention scales quadratically with the sequence length as existing attention models are trained to attend to individual tokens with a fixed granularity,~\eg,~text characters and image pixels. Hence the vanilla Transformer~\cite{NIPS2017_3f5ee243} is prohibitive for learning long sequences such as long documents or high-resolution images (modeled as long sequences of image patches).

Recent works build on the hierarchical architecture to improve the capability of long-range context modeling. In the vision domain for example, hierarchical transformers such as Swin transformer~\cite{liu2021Swin}, PVT~\cite{Wang_2021_ICCV} and ViL~\cite{Zhang_2021_ICCV} rely on predefined image pyramids to compute self-attention at multiple scales, and can thus model long sequences of image patches at a much higher resolution. However, both the scaling scheme and effective attention granularities remain fixed in these methods. In a similar spirit, `area attention'~\cite{pmlr-v97-li19e} computes multi-scale attention which is generic for both language and vision tasks. Specifically, attention is computed against a multi-scale memory formed by pooling the original memory with predetermined pool sizes. This not only requires larger memory but also does not adapt the context range based on content. Finally, the BP-Transformer~\cite{1911-04070} computes attention using multi-scale attention spans that encode fine-to-coarse contexts, but it imposes a sparsity prior on the attention mechanism, which is adaptive, but which might hurt its capacity.

\textbf{Efficient Transformers} mostly use sparsity or low-rank assumptions on the attention matrix to reduce cost. For sparse attention methods, one can sparsify the attention matrix with predefined patterns like~local window~\cite{yang-2018,sukhbaatar2019,child2019sparsetransformer}, blockwise~\cite{blockwise}, log-sparse~\cite{LiJXZCWY19} or axial~\cite{ho2019axial} patterns and their combinations~\cite{Beltagy2020Longformer,ainslie2020etc,zaheer2020bigbird}. The sparsity patterns can also be learned as in~\cite{Kitaev2020Reformer,TACL2405,Tay2020SparseSA}. These sparse attention methods, despite their sub-quadratic cost, often have a reduced model capacity because each token can only attend to a subset of tokens. Generally, sparse attention needs more layers to model full contextual dependencies in a long sequence~\cite{child2019sparsetransformer}. Another family of efficient transformers approximate the attention matrix using low-rank projections~\cite{wang2020linformer} or feature maps of particular kernels~\cite{katharopoulos20}. Such low-rank methods preserve the full attention capability with low computational cost, but suffer from the lossy approximations of potentially full-rank attention. We depart from the above-mentioned methods, aiming for efficient, full attention without sparse or low-rank approximations. Nevertheless, our ContextPool module can be embedded within the internals of several of these models.

There are some recent attempts to accelerate transformers by directly reducing the number of tokens to process in attention layers. Ryoo~\etal~\yrcite{ryoo2021tokenlearner} proposed to `tokenize' the input images by aggregating their feature maps into a few tokens, while DynamicViT~\cite{rao2021dynamicvit} relies on an extra neural network to prune tokens for a fully trained ViT~\cite{dosovitskiy2021an}. We provide a novel perspective for parameter-efficient self-attention given any amount of tokens. By learning to pool the token features with adaptive weighting and pool size, we obtain more expressive tokens from fewer layers.

\textbf{Context in ConvNets} is efficiently captured by convolutions, which summarize local neighborhoods with shared weights and when combined with pooling, can model long-term dependencies. Recent works indicate that ConvNets benefit from using different kernel sizes at different convolutional layers~\cite{Pintea21}. Therefore, many methods choose to learn adaptive kernel size to account for data-dependent context or receptive field. Concretely, they scale kernels by dilation and learn dilation factors over shifted Delta-dirac functions~\cite{8237351}, scalable Gaussian functions~\cite{Shelhamer19} or Gaussian derivative filters~\cite{Pintea21,tomen21a}. Another method of receptive field learning in ConvNets is based on learning pooling functions with adaptive pooling regions~\cite{NIPS2011_6c1da886,6248076}. Our ContextPool method is also applicable to ConvNets. By learning dynamic pooling weights and support size, it is shown to be competitive with existing methods while maintaining low computational cost.

\begin{figure*}[!t]
\vskip 0.1in
\begin{center}
\centerline{\includegraphics[width=\linewidth]{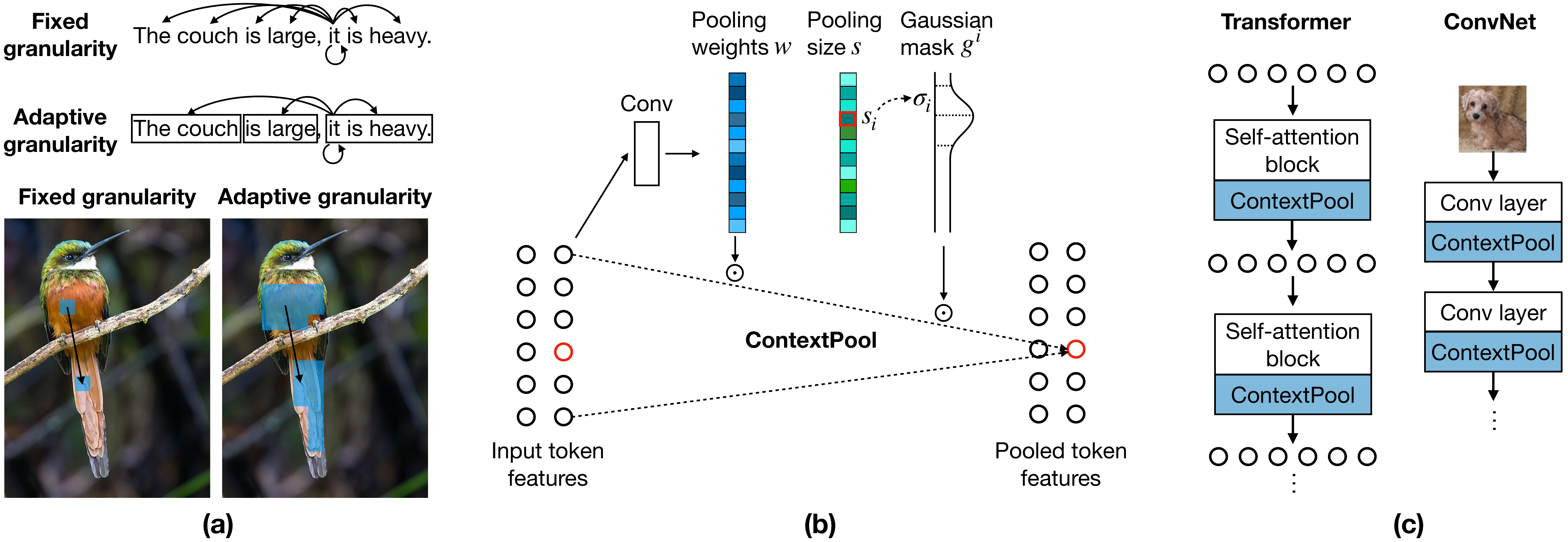}}
\vskip -0.05in
\caption{(a) Motivation: the proposed ContextPool seeks to achieve adaptive attention granularity through adaptive context pooling around each token and then computing context-wise attention. This helps to capture high-level dependencies and is useful to model ambiguous pronoun ``\emph{it}'' by associating with neighboring phrases rather than single words, or to model interactions between varying-sized object parts. (b) For adaptive ContextPool, we learn the pooling weights and support size dynamically for each token. (c) Our ContextPool module is applicable to both transformers and ConvNets for efficient feature learning. For transformers, the ContextPool module is placed after each attention block, whose output token features are pooled to the same number of features for use in the next attention block. While for ConvNets, ContextPool replaces the conventional pooling function (please refer to supplementary materials for details).}
\label{fig:schematic}
\end{center}
\vskip -0.2in
\end{figure*}

\section{ContextPool for Transformers}

\subsection{Standard Transformers}

A standard transformer model~\cite{NIPS2017_3f5ee243} is a chain of self-attention modules (self-attention plus feed-forward layers). The input of each self-attention layer is a feature matrix $\bm{X} \in \mathbb{R}^{n \times d}$ from the preceding layer. $\bm{X}$ is a sequence of $n$ tokens =$\{\bm{x}_1,\dots,\bm{x}_n\}$ each of dimension $d$. The attention layer operates on all the token features in $\bm{X}$. Specifically, each token $\bm{x}_i$ is first transformed to the query $\bm{q}_i=\bm{W}^q \bm{x}_i$, key $\bm{k}_i=\bm{W}^k \bm{x}_i$ and value $\bm{v}_i=\bm{W}^v \bm{x}_i$ with learned projection matrices $\{\bm{W}^q, \bm{W}^k, \bm{W}^v\} \in \mathbb{R}^{d \times d}$. Then the attention score of one query $\bm{q}$ attending to all the keys $\{\bm{k}_i\}$ stored in a memory is given by: 
\begin{equation}
a_i = \frac{\exp({\bm{q}^T\bm{k}_i})}{\sum_{j=1}^n \exp({\bm{q}^T\bm{k}_j})}.
\label{eq1}
\end{equation}
The final output $\bm{o}_q$ from querying the memory with $\bm{q}$ is obtained by taking a weighted average of all the values $\{\bm{v}_i\}$ in memory:
\begin{equation}
\bm{o}_q = \sum_{i=1}^n a_i \bm{v}_i.
\label{eq2}
\end{equation}
In practice, multi-head self-attention is used in transformers, where multiple projections are learned to compute attention within different heads. The outputs are then concatenated and projected into refined token features.

\textbf{Drawback} The above self-attention mechanism assumes a fixed granularity over which to construct the query and key vectors for individual tokens. However, such a fixed granularity may be sub-optimal for modeling context with different scales. Consider neural machine translation with word-based tokens -- translating numerals from one language to another requires little context, but translating ambiguous pronouns (\eg,~``\emph{it}'') requires long range-contextual cues from neighboring tokens. 
One might argue that this difficulty can be resolved by using deeper models, where self-attention in deeper layers can capture the interactions between single tokens and bake them into deep feature representations. This can progressively correct for the fixed attention granularity at lower layers, but requires more computation that may be avoidable with an adaptive strategy.

\subsection{Adaptive Context Pooling}

\textbf{Motivation} We motivate our method using Fig.~\ref{fig:schematic}(a). In language modelling, if we can piece words together to form phrases, we can gradually capture the phrasal patterns and useful context information. This helps to disambiguate the pronoun ``\emph{it}'' by linking it to the phrase ``\emph{The couch}''. Similarly, in image understanding, pooling similar image patches can enable the model to learn semantics of a bird's body parts. To account for the special role of context in obtaining adaptive attention granularity, we introduce an explicit way of learning context-aware token features. We do this by learning to pool neighboring features for each token (\emph{ContextPool}). Self-attention between such pooled features can thus be context-aware and model high-level dependencies, without requiring multiple self-attention layers.  

Therefore, our ContextPool method needs an input-adaptive pooling function. Below, we describe how to learn that with adaptive pooling weights and pooling size. Specifically, given the input token feature matrix $\bm{X} \in \mathbb{R}^{n \times d}$, we pool for each token $\bm{x}_i \in \bm{X}$ with learned weights $\bm{w}  \in \mathbb{R}^{n \times 1}$ and a Gaussian mask $\bm{g}^i  \in \mathbb{R}^{n \times 1}$ (acting as a soft, local pooling window), generating a contextual feature matrix $\bm{Y} \in \mathbb{R}^{n \times d}$ of the same size of $\bm{X}$ (see Fig.~\ref{fig:schematic}(b)).

\textbf{Adaptive pooling weights} differ from the uniform ones in the popular average pooling function. We found it helpful to reweight the neighboring token features $\{\bm{x}_j\}$ during pooling based on their contextual support to $\bm{x}_i$. One widely used approach of measuring such support is based on nonlocal feature similarity as in~\cite{8578911}:
\begin{equation}
w_j = \frac{\exp(\theta(\bm{x}_i)^T \phi(\bm{x}_j))}{\sum_{j=1}^n \exp(\theta(\bm{x}_i)^T \phi(\bm{x}_j))},
\label{eq3}
\end{equation}
where $\theta(\bm{x}_i)=\bm{W}^{\theta} \bm{x}_i$ and $\phi(\bm{x}_j)= \bm{W}^{\phi} \bm{x}_j$ are embeddings with learnable projections $\{\bm{W}^{\theta},\bm{W}^{\phi}\} \in \mathbb{R}^{d \times d}$.

We dub such learned pooling weights $\bm{w}$ as nonlocal weights (\textbf{NL weights}). The intuition behind NL weights is that similar features in the context are likely to correspond to semantically related entities. Therefore, nonlocal similarity pooling in form of $\sum_{i=1}^n w_i \bm{x}_i$ can provide contextual information to increase (or decrease) the probability of a semantic region or segment. Note we only introduce NL weights as a comparing baseline.

One limitation of NL weights is that each weight $w_j$ in Eq.~(\ref{eq3}) only depends on a feature pair $(\bm{x}_i,\bm{x}_j)$, overlooking the potential contributions from other features to $\bm{x}_i$. Here we turn to learning $w_j$ by a mapping function $m(\cdot)$ conditioned on all the token features $\{\bm{x}_i\}$ in $\bm{X}$. In fact, we predict the pooling weights $\bm{w}=m(\bm{X})$ all at once, where $m(\cdot)$ is implemented as two convolutional layers. Hence the prediction of $\bm{w}$ is collaborative and more efficient than NL weights prediction.

\textbf{Adaptive pooling size} Pooling with adaptive weights, however, does not take into account the location relationships between tokens. Here we introduce a locality prior to bias pooling towards the local context around considered token. Note that learning the pooling weights alone might also be able to find local patterns in the learned weights. However, the locality prior can simplify learning by allowing factorized and independent predictions of pooling weights and scope. Our experiments support this hypothesis with favorable results. The locality prior also shares a similar high-level idea with the effective receptive field~\cite{NIPS2016_c8067ad1}, which is shown to have a Gaussian distribution.

We learn a Gaussian mask for each token to implement soft, localized pooling with \emph{adaptive} pooling size rather than a hand-picked one. Specifically, we learn the mapping function $m(\cdot)$ to predict both the pooling weights $\bm{w}\in \mathbb{R}^{n \times 1}$ and sizes $\bm{s}\in \mathbb{R}^{n \times 1}$ for $n$ input tokens conditioned on their features $\bm{X}$,~\ie,~$\{\bm{w},\bm{s}\}=m(\bm{X})$. We implement $m(\cdot)$ again by two convolutional layers, but with the channel size set to 2 now. This enables generating the vectors of $\bm{w}$ and $\bm{s}$ altogether, which are normalized by a softmax function for ease of training. Given the normalized pooling size $s_{i} \in [0,1]$, we then transform it to the standard deviation $\sigma_{i}=rn \cdot s_{i}$ of a Gaussian mask $\bm{g}^i \sim \mathcal{N}(i,\sigma_{i}^2)$. Here $r$ is a scalar empirically set as 0.1.

By multiplying the learned pooling weights $\bm{w}$ with the Gaussian mask $\bm{g}^i$ for token $\bm{x}_i$, we arrive at our final ContextPool function:
\begin{equation}
\bm{y}_i = f_{ave}(\bm{X} \odot \gamma(\bm{w}) \odot \gamma(\bm{g}^i)) = \sum_{j=1}^n \bm{x}_{j} \cdot w_{j} \cdot g^i_{j},
\label{eq4}
\end{equation}
where $\bm{y}_i \in \bm{Y}$ denotes the ContextPooled features, $f_{ave}$ denotes average pooling function, $\gamma(\cdot)$ is a broadcasting function for element-wise multiplication $\odot$. We set the normalization factor as $C(\bm{X}) = \sum_{j} w_{j} \cdot g^i_{j}$.

As shown in Fig.~\ref{fig:schematic}(c), our ContextPool module can be placed after different attention blocks. After each attention block, we take its outputs as input token features and pool them to the same number of features for use in the next attention block. During training, we jointly learn the main model and ContextPool parameters. We also show the applicability of our ContextPool method to ConvNets in supplementary materials.

\section{Results}

We evaluate the proposed ContextPool (dubbed ``CP'' as a prefix) module mainly in the transformer architecture to show how strengthened context modeling can benefit self-attention in a parameter-efficient way. We validate such benefits on both language and vision tasks that require a good context modeling capability. Supplementary materials also show that our ContextPool can be seamlessly integrated into ConvNets in place of the conventional pooling function. ContextPool leads to strong results on standard image classification benchmarks, being competitive or even better than those ConvNets with adaptive kernel size or receptive field. This comes at low computational overhead, showing the potential of ContextPool to be a generic module for efficient representation learning.

\subsection{Tasks, Datasets and Implementation}

\textbf{Neural Machine Translation} For language tasks, we first experiment on the token-level Neural Machine Translation (NMT) task. We use both the WMT 2014 English-to-German (EN-DE) dataset with about 4.5 million English-German sentence pairs, and the and English-French (EN-FR) dataset with about 36 million English-French sentence pairs. A token is a byte pair or a word piece as in~\cite{NIPS2017_3f5ee243}. We compare with different methods all using three transformer architectures as defined in~\cite{pmlr-v97-li19e}: Small (2 layers), Base and Big (6 layers) models. For our method, we insert ContextPool after every attention layer. Following~\cite{pmlr-v97-li19e}, we train for 250k iterations for Small and Base models, and for 600k iterations with a smaller batch size for the Big model due to the memory constraint. We use Adam optimizer with the same learning rate schedule in~\cite{NIPS2017_3f5ee243}.

\textbf{Autoregressive Language Modeling} We also evaluate ContextPool on the autoregressive language modeling task at character level. Compared to the token-level task, character-level task is harder due to much longer sequences, which would hypothetically benefit more from stronger context modeling. We use enwik8 and text8 datasets, each with 100M characters and 90M/5M/5M for train/dev/test as in~\cite{MattMahoney09}. For testing, we follow~\cite{Beltagy2020Longformer} to split the dataset into overlapping sequences of length 32k with step size 512, and then calculate the Bits Per Character (BPC) of predicting 512 characters from previous 32k.

We use the same 12-layer model architecture with Longformer~\cite{Beltagy2020Longformer}. We train our models in 3 stages with increasing sequence lengths (2048, 4096, 8192) and different batch sizes (32, 32, 16). All models are trained for a total of 530k steps with linear learning rate warmup. We also use dropout rates 0.2 and weight decays 0.01.

\textbf{Image Classification} We benchmark different transformer models on the widely used ImageNet-1K classification dataset~\cite{DenDon09Imagenet}. There are 1.28M training and 50k validation images from 1k classes. The top-1 accuracy on a single crop
is reported. We consider the regular training setting in~\cite{touvron21a} where no external training data are used. The input image resolution is $224^2$ by default. For higher resolutions like $384^2$, we fine-tune the $224^2$ trained models. We train for 300 epochs with the AdamW optimizer, using a cosine decay learning rate scheduler and linear warm-up (20 epochs). When fine-tuning on higher resolution images, we tune for 30 epochs with a similar training recipe as in~\cite{liu2021Swin}. We have batch size 1024, initial learning rate 0.001, weight decay 0.05, and the max norm of gradient clipping 1. Stronger data augmentation is found to benefit our ContextPool method. Therefore we use a larger degree of augmentation with the augmentation techniques in~\cite{touvron21a} such as RandAugment~\cite{NEURIPS2020_d85b63ef}, making our pooled token features more robust.

\subsection{Ablations and Comparisons}

\textbf{Ablation on adaptive pooling weights and size} We start with ablation studies on these two core components of our ContextPool method and compare against their alternatives. For this purpose, both the NMT and image classification tasks are considered for a comprehensive comparison. For NMT, we choose the English-German (EN-DE) translation task using the Base model. While for image classification, the ViT-B/16 model~\cite{dosovitskiy2021an} (the ``Base'' variant with $16\times16$ input patch size) is used.

\begin{table*}[!t]

\begin{minipage}{.5\textwidth}
\caption{Ablations on token-level translation (EN-DE task) using the Base model. Speed (steps / s) is measured on a V100 GPU. CP denotes the use of our full ContextPool module ($\bm{w} \odot \bm{g}^i$). The middle and bottom cells compare with alternative weightings and locality priors respectively for context pooling.}
\label{tb:ablation_NMT}
\begin{center}
\begin{small}
\resizebox{\linewidth}{!}{
\begin{tabular}{lccc}
\toprule
Method & Memory (G)& Speed & BLEU $\uparrow$ \\
\midrule
Base & 17.2 & 1.20 & 28.16\\
CP-Base ($\bm{w} \odot \bm{g}^i$)& 17.6 & 1.12 & \textbf{28.91}\\ \midrule
Unnormalized weights $\odot \,\bm{g}^i$ & 17.6 & 1.13 & 28.79\\
Uniform weights $\odot\, \bm{g}^i$  & 17.4 & 1.16 & 28.52\\
NL weights $\odot\, \bm{g}^i$ & 21.3 & 0.84 & 28.66\\ \midrule
No locality prior $\odot\, \bm{w}$ & 17.4 & 1.15 & 28.31\\
Fixed window $\odot\, \bm{w}$& 17.4 & 1.15 & 28.55\\
Adaptive window $\odot\, \bm{w}$& 17.6 & 1.12 & 28.74\\
Random sparse $\odot\, \bm{w}$& 17.4 & 1.15 & 28.14\\
\bottomrule
\end{tabular}
}
\end{small}
\end{center}
\vskip -0.2in
\end{minipage}
\hspace{0.1in}
\begin{minipage}{.5\textwidth}
\caption{Ablations on ImageNet-1K classification. Top1 is top-1 accuracy. Throughput (images / s) is measured on a V100 GPU. CP denotes the use of our full ContextPool module ($\bm{w} \odot \bm{g}^i$). The middle and bottom cells compare with alternative weightings and locality priors respectively for context pooling.}
\label{tb:ablation_imagenet}
\begin{center}
\begin{small}
\resizebox{\linewidth}{!}{
\begin{tabular}{lccc}
\toprule
Method & FLOPs (G) & Throughput & Top1 $\uparrow$ \\
\midrule
ViT-B/16 & 55.4 & 85.9 & 77.9\\
CP-ViT-B/16 ($\bm{w} \odot \bm{g}^i$) & 56.7 & 84.1 & \textbf{79.9}\\ \midrule
Unnormalized weights $\odot \, \bm{g}^i$ & 56.6 & 84.2 & 79.7\\
Uniform weights $\odot \, \bm{g}^i$ & 56.1 & 84.8 & 78.9\\
NL weights $\odot \, \bm{g}^i$ & 68.8 & 69.2 & 79.4\\ \midrule
No locality prior $\odot \, \bm{w}$& 56.0 & 85.1 & 78.3\\
Fixed window $\odot \, \bm{w}$& 56.0 & 85.1 & 78.9\\
Adaptive window $\odot \, \bm{w}$& 56.7 & 84.1 & 79.6\\
Random sparse $\odot \, \bm{w}$& 56.0 & 85.1 & 78.1\\
\bottomrule
\end{tabular}
}
\end{small}
\end{center}
\vskip -0.2in
\end{minipage}
\vskip -0.2in
\end{table*}

Tables~\ref{tb:ablation_NMT} and~\ref{tb:ablation_imagenet} summarize the results. We observe that our ContextPool method can consistently improve the baseline transformers at only marginal overhead (in memory, FLOPs and speed), due to the efficiency of adaptive pooling functions implemented by convolutions. For the learning of pooling weights, we first compare with those un-normalized weights without using softmax (middle cell). We obtained slightly worse results for both tasks using un-normalized weights, which confirms the need of normalization for effective weighting (note the pooling size predictions were always softmax normalized).
One straightforward alternative to our learned weighting is the use of uniform weights,~\ie,~to perform average pooling. By doing so, we save the learning cost for the weights but suffer from apparent performance loss. We can also choose to learn NL weights as in Eq.~(\ref{eq3}), which is equivalent to learning extra, single-head self-attention weights in transformers and is thus much more costly than our lightweight convolutional method. Further, NL weights are found to be less competitive than ours due to the lack of feature interactions in pairwise weights computation.

\begin{table*}[!t]
\caption{Ablation study on the design choice of ContextPool (CP) module. For transformers, we choose the same NMT and image classification tasks as in Tables~\ref{tb:ablation_NMT} and~\ref{tb:ablation_imagenet}, with identical task settings and baseline models. We also include the ConvNet experiments (details in supplementary materials) for more comprehensive ablations.}
\label{tb:CP_design}
\begin{center}
\begin{small}
\resizebox{\textwidth}{!}{
\begin{tabular}{cccccccccc}
\toprule
\multirow{2}{*}{Method} & \multicolumn{3}{c}{\textbf{Transformer} (EN-DE translation)} & \multicolumn{3}{c}{\textbf{Transformer} (ImageNet classification)} & \multicolumn{3}{c}{\textbf{ConvNet} (CIFAR-10 classification)} \\
\cmidrule(lr){2-4}
\cmidrule(lr){5-7}
\cmidrule(lr){8-10}
 & Memory (G)& Speed & BLEU $\uparrow$ & FLOPs (G) & Throughput & Top1 $\uparrow$ & FLOPs (G) & Size & Accuracy $\uparrow$ \\
\midrule
Baseline & 17.2 & 1.20 & 28.16 & 55.4 & 85.9 & 77.9 & 3.7 & 0.66M & 92.9 \\
+ CNN-based CP (default) & 17.6 & 1.12 & \textbf{28.91} & 56.7 & 84.1 & \textbf{79.9} & 3.9 & 0.68M& \textbf{93.4} \\
+ MLP-based CP & 17.3 & 1.17 & 28.33 & 56.5 & 85.2 & 78.7 & 3.8 & 0.67M& 93.1 \\
+ Self-attention-based CP & 21.3 & 0.84 & 28.66 & 68.8 & 69.2 & 79.4 & 4.4 & 0.67M& 93.2 \\
\bottomrule
\end{tabular}
}
\end{small}
\end{center}
\vskip -0.1in
\end{table*}

Tables~\ref{tb:ablation_NMT} and~\ref{tb:ablation_imagenet} (bottom cell) compare several baselines to replace our learned Gaussian mask that imposes a soft locality prior for pooling. When we remove the locality prior entirely, we save compute again but observe a big drop in performance for both tasks. This suggests that context pooling indeed benefits from a local ``receptive field'' (similar to the findings in~\cite{NIPS2016_c8067ad1}). It also suggests the difficulty of disentangling the local prior from the pooling weights by learning the latter alone in an unfactorized way. The ``Fixed window'' baseline is one simple remedy to this issue by associating a fixed local window to the pooling function, where the window size is hand-picked on validation data. We see immediate help from this baseline (relative to ``no locality prior''). On the other hand, we find pooling at random sparse locations will slightly hurt performance. Finally, learning adaptive local windows performs close to our method with adaptive soft Gaussian masks, but the benefits of the latter still hold with consistent gains.

\textbf{Ablation on the design choice of ContextPool module} Recall that our default ContextPool module is implemented as a convolutional mapping function $m(\bm{X})$, which maps the input feature matrix $\bm{X}$ into arrays of pooling weights and size. Table~\ref{tb:CP_design} compares such a CNN-based design choice against alternatives like fully-connected MLP and self-attention layers. Here we conduct the comparing experiments on both transformers and ConvNets (detailed in supplementary materials) for a more comprehensive ablation. Note for transformers, we still benchmark on the same tasks as in Tables~\ref{tb:ablation_NMT} and~\ref{tb:ablation_imagenet}, with identical task settings and baseline models.

The MLP-based ContextPool module in Table~\ref{tb:CP_design} can be considered as the simplest form of $m(\cdot)$, which maps each feature vector $\bm{x}_i \in \bm{X}$ to its corresponding pooling weights. We can see that MLP is more compute-efficient than our convolutional module but worse in performance. The reason is that such MLP module operates individually for $\bm{x}_i$ without considering feature interactions when predicting their pooling weights, while convolutional layers leverage neighboring features to do so. Note we can use a giant MLP that predicts for all $\{\bm{x}_i\}$ together, which becomes collaborative but at a much higher cost.

Alternatively, we can implement $m(\cdot)$ using a (single-head) self-attention layer as in Eq.~(\ref{eq3}). However, as mentioned before, such mapping function $m(\cdot)$ is not only costly with quadratic computation, but also limited in modeling feature interactions. As shown in Table~\ref{tb:CP_design}, the inferiority also translates to the ConvNet framework. Note we can improve by modeling richer feature interactions with more than one attention layers, but this will further increase the cost.

\begin{figure}[!t]
\begin{center}
\centerline{\includegraphics[width=0.7\columnwidth]{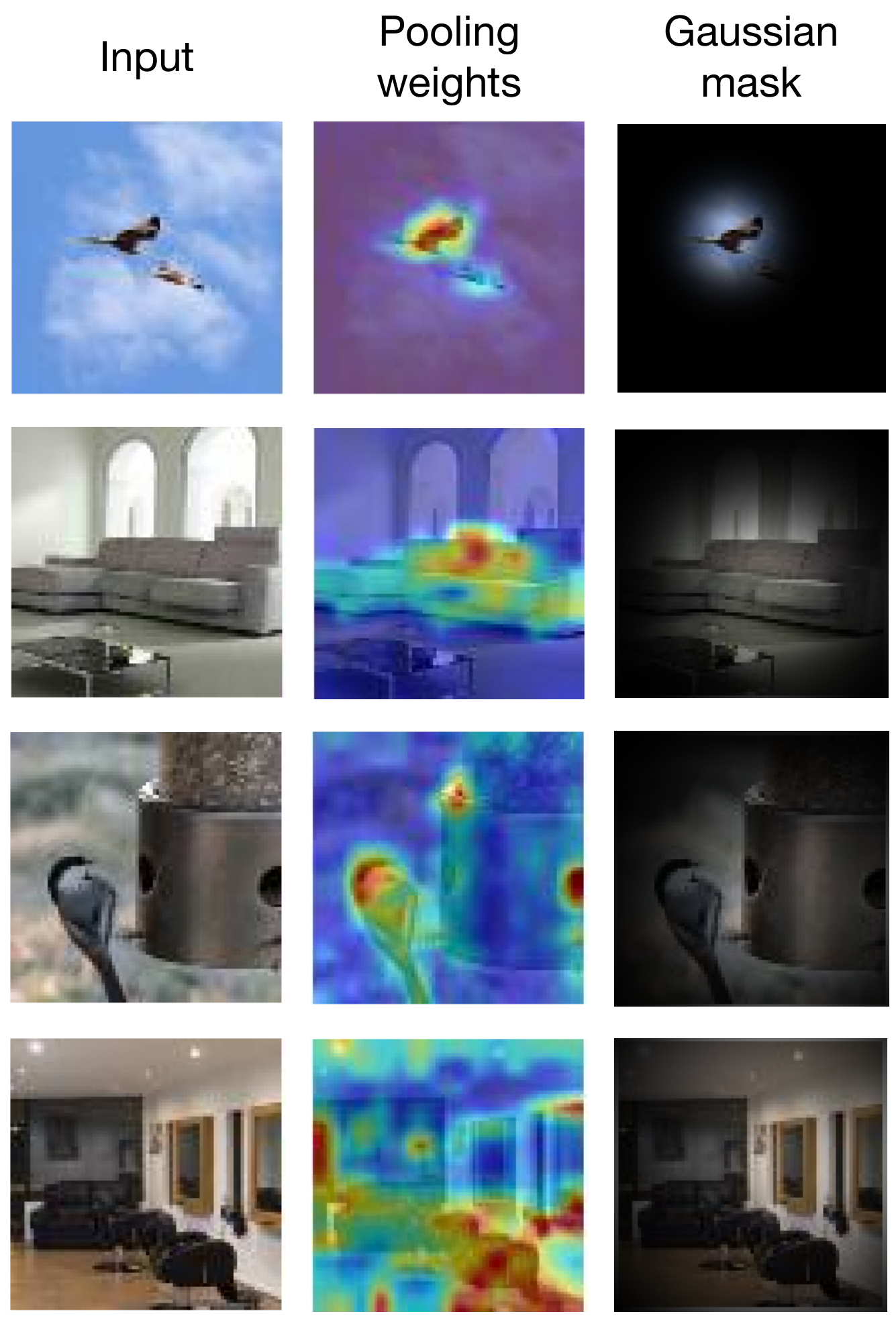}}
\vskip -0.1in
\caption{Visualizations of the pooling weights and size (in the form of soft Gaussian mask) predicted by our ContextPool module on example ImageNet images. We observe that the pooling weights are learned to aggregate diverse information from different locations or object parts, while the pooling size is learned to capture either local or global image context depending on the input.}
\label{fig:visualization}
\end{center}
\vskip -0.3in
\end{figure}

\begin{table*}[!t]
\caption{The BLEU scores for token-level translation on the WMT 2014 EN-DE and EN-FR datasets. We compare our CP-attention with standard attention~\cite{NIPS2017_3f5ee243}, local attention~\cite{yang-2018} and area attention~\cite{pmlr-v97-li19e}.}
\vskip 0.1in
\label{tb:NMT}
\begin{center}
\begin{small}
\begin{tabular}{ccccccccc}
\toprule
\multirow{2}{*}{Model} & \multicolumn{2}{c}{Standard attention} & \multicolumn{2}{c}{Local attention} & \multicolumn{2}{c}{Area attention} & \multicolumn{2}{c}{CP-attention (ours)}\\
\cmidrule(lr){2-3}
\cmidrule(lr){4-5}
\cmidrule(lr){6-7}
\cmidrule(lr){8-9}
 & EN-DE & EN-FR & EN-DE & EN-FR & EN-DE & EN-FR & EN-DE & EN-FR\\
\midrule
Small & 22.55 & 31.93 & 22.71  & 32.48  & 23.20 & 32.93 &  \textbf{23.67} & \textbf{33.24}  \\
Base & 28.16 & 38.97 & 28.32  &  39.04  & 28.52 & 39.19 & \textbf{28.91}  &  \textbf{39.36} \\
Big & 29.26 & 41.00 & 29.31  &  41.17 & 29.77 & 41.46 &  \textbf{30.11} &  \textbf{41.59} \\
\bottomrule
\end{tabular}
\end{small}
\end{center}
\vskip -0.1in
\end{table*}

\subsection{Visualizations and Analysis}

Now we visualize what have been learned in our pooling weights and pooling sizes (in the form of soft Gaussian mask). Since visualization is easier on images with spatial grids, we take the ViT-B/16 model and visualize the predictions from our ContextPool module after the second attention layer.

\begin{figure}[!t]
\begin{center}
\centerline{\includegraphics[width=\columnwidth]{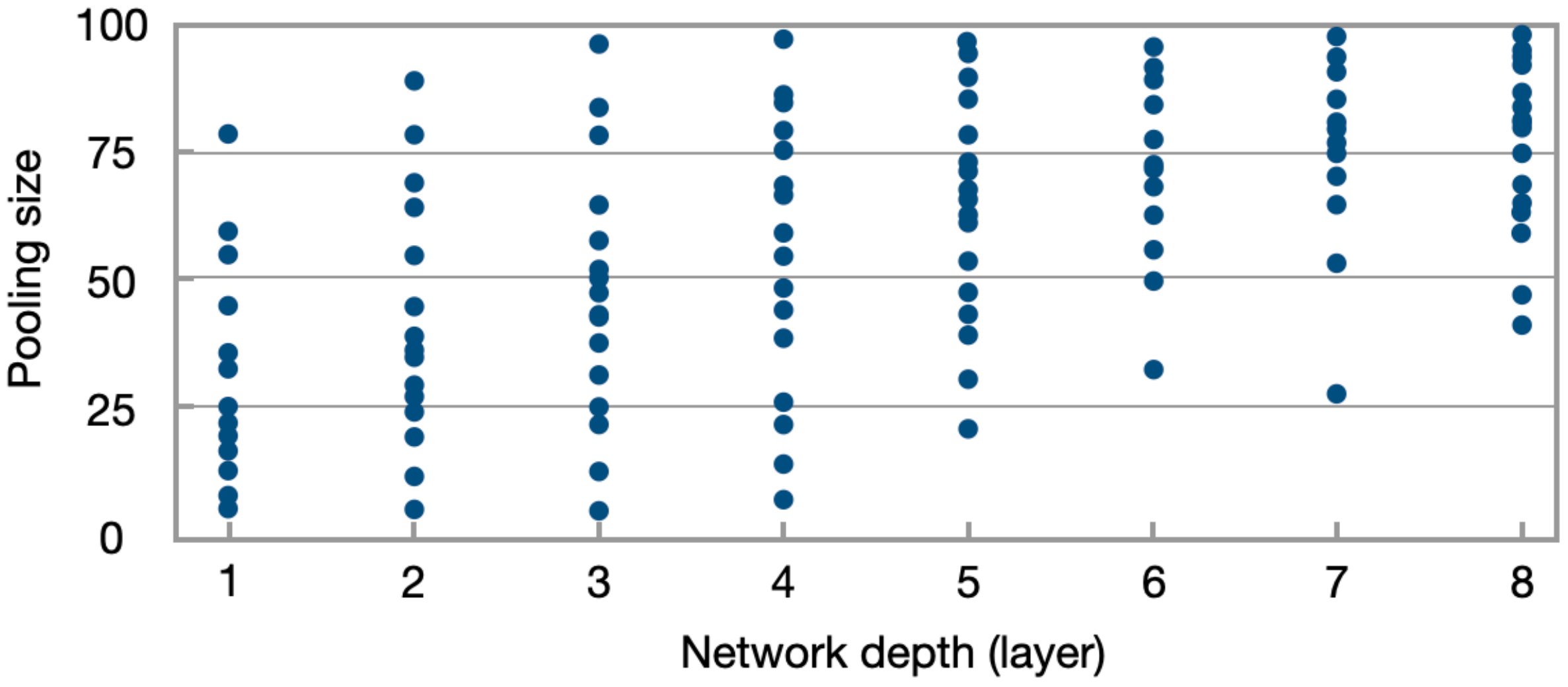}}
\vskip -0.1in
\caption{Distributions of the predicted pooling size by our ContextPool module in different attention layers (ViT-B/16).}
\label{fig:Poolsize_dist}
\end{center}
\vskip -0.3in
\end{figure}

We are able to observe from Fig.~\ref{fig:visualization} that: 1) The pooling weights are learned to aggregate diverse information, and seem to go beyond feature similarity (the main intuition of NL weights). The last image gives one example where the pooling weights highlight some dissimilar regions around the window and ceiling, which can instead accumulate evidence for the target class of ``room''. 2) The learned pooling size is indeed input dependent, capturing the local or global context adaptively. Fig.~\ref{fig:Poolsize_dist} further shows the distributions of pooling size in different layers. Interestingly, the predicted pooling size remains diverse within each layer, but in general tends to increase at higher layers to capture long-range dependencies.

\subsection{Comparing to SOTAs on Language Tasks}

Table~\ref{tb:NMT} evaluates our ContextPool-based attention model on the token-level NMT task using both EN-DE and EN-FR datasets. Comparison is made against standard attention and other variants that model context differently. Three transformer architectures are adopted as in~\cite{pmlr-v97-li19e}.

It is observed that local attention only achieves marginal gains over standard attention, mainly because the locality is added to the attention mechanism which hurts the full attention capacity. Area attention preserves full attention by allowing queries to attend to the whole memory. The memory is a multi-scale one to encode context of varying scales. Despite the strong BLEU scores from area attention, it is not flexible enough to model content-dependent context due to the use of fixed set of pooling sizes when constructing the multi-scale memory. Our ContextPool is able to meaningfully outperform area attention across datasets and model sizes, thanks to its adaptiveness during context pooling.

\begin{figure}[!t]
\begin{center}
\centerline{\includegraphics[width=0.85\columnwidth]{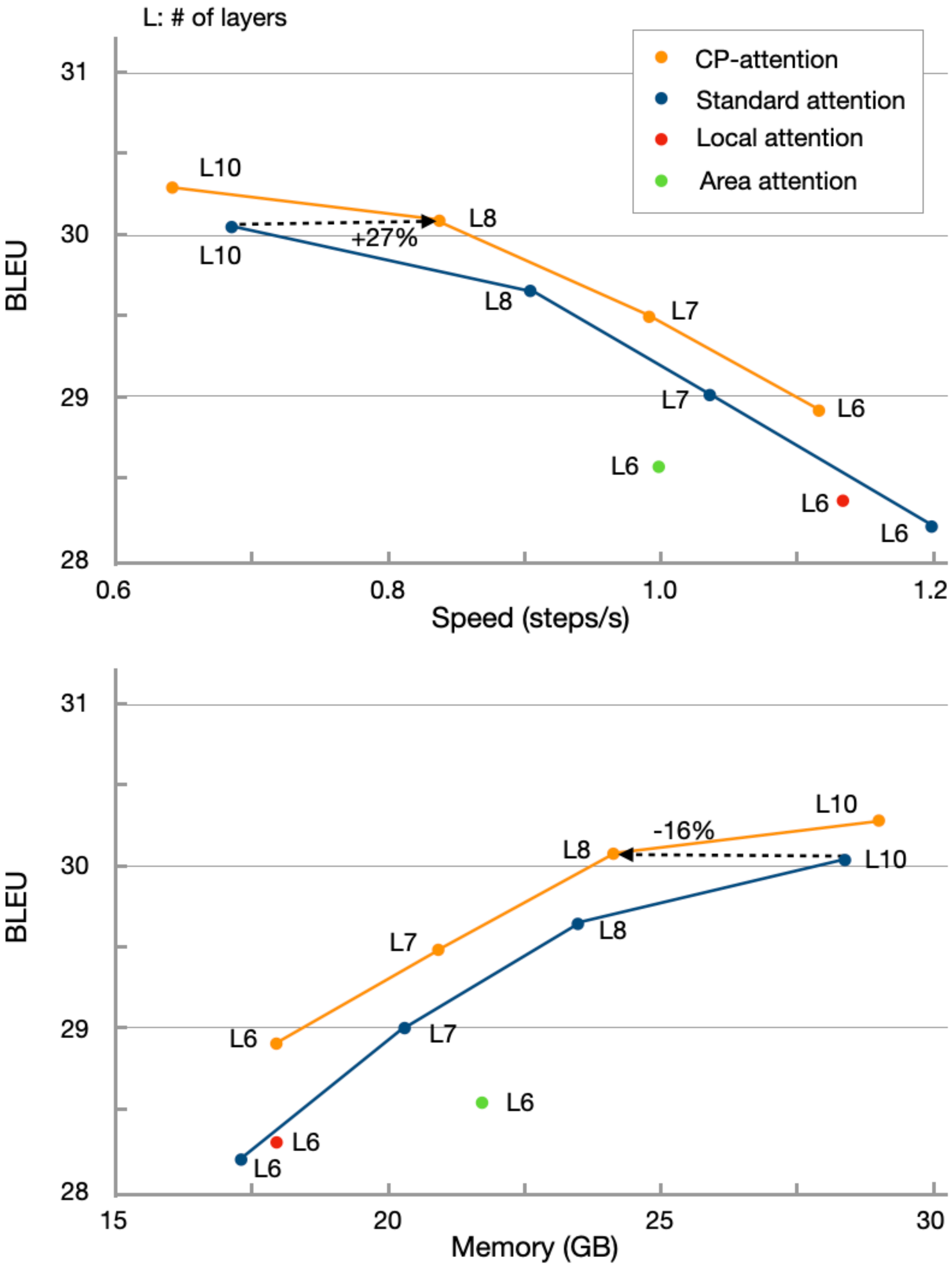}}
\caption{Performance-cost comparisons on token-level translation (EN-DE task) using the Base model. The number of layers ranges from $L=6$ to 10.}
\label{fig:NMT}
\end{center}
\vskip -0.3in
\end{figure}

Fig.~\ref{fig:NMT} further compares the above methods in terms of computation and memory complexities. Given the default number of layers $L=6$, our CP-attention not only outperforms others at the same $L$, but also strikes a better trade-off between performance and cost. For instance, our CP-attention ($L=6$) obtains a higher BLEU score 28.91 at a noticeably faster speed and lower memory than area attention, since the latter needs to maintain a multi-scale memory online. More importantly, we are able to utilize our saved compute in the form of additional layers (increasing $L$ to 7). This way, we further improve the model capacity and BLEU score, but our speed and memory remain comparable to those of area attention with only $L=6$ layers. When we continue to train a deeper model with CP, we found significantly boosted parameter efficiency over the one without CP. Interestingly, our CP-attention with $L=8$ layers obtains a similar BLEU score with the 10-layer vanilla attention model (without CP), leading to $27\%$ faster speed and $16\%$ less memory used. On the other hand, when we train shallower models, the performance gap (with vs. without CP) becomes larger,~\eg,~$\Delta$BLEU$=$1.21 when $L=$4. This again demonstrates our improved model expressiveness.

\begin{table}[t]
\caption{BPC ($\downarrow$) and model size on enwik8 and text8 for autoregressive language modeling. The number of layers is included in the parenthesis. CP denotes the use of our ContextPool module.}
\label{tb:autoregressive}
\begin{center}
\begin{small}
\begin{tabular}{cccccc}
\toprule
\multirow{2}{*}{Model} & \multirow{2}{*}{\#Param} & \multicolumn{2}{c}{text8} & \multicolumn{2}{c}{enwik8}\\
\cmidrule(lr){3-4}
\cmidrule(lr){5-6}
 & & Dev & Test & Dev & Test\\
\midrule
T12 & 44M & - & 1.18  & -  & 1.11\\
Transformer-XL & 41M & - & -  & -  & 1.06\\
Adaptive local & 38M & 1.05 & 1.11  & 1.04  & 1.02\\
BP-Transformer & 38M & - & 1.11  & -  & 1.02\\
Longformer & 41M & 1.04 & 1.10  & 1.02  & 1.00\\
Reformer & - & - & -  & -  & 1.05\\ \midrule
CP-Transformer (12) & 39M & 1.04 & 1.09  & 1.02  & 0.99\\
CP-Transformer (14) & 44M & \textbf{1.02} & \textbf{1.07}  & \textbf{1.01}  & \textbf{0.97}\\
CP-Transformer (11) & \textbf{36M} & 1.05 & 1.11  & 1.03  & 1.01\\ \midrule
CP-Adaptive local & 39M & 1.05 & 1.10  & 1.03  & 1.01\\
CP-Longformer & 43M & 1.03 & 1.09  & 1.02  & 0.99\\
\bottomrule
\end{tabular}
\end{small}
\end{center}
\vskip -0.2in
\end{table}


Finally, we evaluate on the challenging task of character-level autoregressive language modeling (see Table~\ref{tb:autoregressive}). BPC results are reported on the Dev/Test sets of enwik8 and text8 datasets. We compare with the baseline models of T12~\cite{RfouCCGJ19} and Transformer-XL~\cite{dai-2019-transformer}, as well as four representative methods of sparse attention. Among them, Adaptive local~\cite{sukhbaatar2019} and BP-Transformer~\cite{1911-04070} use the local window as a sparsity prior, but with a learned window size and multi-scale windows respectively. Longformer~\cite{Beltagy2020Longformer} uses a combined sparsity pattern (global+local window), while Reformer~\cite{Kitaev2020Reformer} chooses to learn the patterns.

The above sparse attention methods differ from our ContextPool method in their loss of full attention capacity despite the improved efficiency. Our method on the other hand, computes full attention over ContextPooled token features. Note our context pooling function does have a locality prior, similar to existing sparsity priors based on local window. But the critical difference is that our locality prior is only applied to feature pooling, not to the following full attention process.

Table~\ref{tb:autoregressive} confirms the benefits of full attention models. Our CP module when applied to the standard 12-layer transformer, makes a strong baseline CP-Transformer (12) that has a small model size (39M parameters) but consistently outperforms the compared sparse attention methods. We are able to further lower the model size to 36M when inserting CP to a 11-layer model without sacrificing the performance much, due to the boosted model expressiveness. When the saved parameters are re-invested in constructing a deeper model (14 layers) that has comparable model size of 44M, we attain new state-of-the-art performance on both enwik8 and text8.

The bottom cell of Table~\ref{tb:autoregressive} examines if our ContextPool is complementary to sparse attention. The answer is positive given our consistent gains over two sparse attention baselines. Intuitively, sparse attention would benefit more from our expressive token features that are context-aware.

\subsection{Comparing to SOTAs on Image Classification}
Table~\ref{tb:imagenet} evaluates our CP method on ImageNet classification and compares with the state-of-the-art methods ViT~\cite{dosovitskiy2021an}, DeiT~\cite{touvron21a} and Swin-T~\cite{liu2021Swin}. It is shown that when CP is simply applied to the 12-layer ViT-B/16 model, performance gains are achieved at low overhead. When we plug CP into a smaller CP-ViT-B/16 model with 10 layers, this model can perform even comparably to ViT-L/16 despite being much more efficient. We further show CP is applicable to the Swin transformer that computes multi-scale attention with an image pyramid. Our CP method proves helpful for the two Swin-B models using different input image resolutions, and achieves a strong top1 accuracy of 85.6\%.

\begin{table}[t]
\caption{ImageNet-1K top1 classification accuracy. Throughput (images/s) is measured on a V100 GPU. CP denotes the use of our ContextPool module. The number of attention layers is included in the parenthesis.}
\label{tb:imagenet}
\begin{center}
\begin{small}
\resizebox{1.05\linewidth}{!}{
\begin{tabular}{cccccc}
\toprule
Method & image & \#Param & FLOPS & image/s & Top1 \\
\midrule
ViT-B/16 & $384^2$ & 86M & 55.4G & 85.9 & 77.9\\
ViT-L/16 & $384^2$ & 307M& 190.7G & 27.3 & 76.5\\\midrule
DeiT-S & $224^2$ & 22M & 4.6G & 940.4 & 79.8\\
DeiT-B & $224^2$ & 86M & 17.5G & 292.3 & 81.8\\
DeiT-B & $384^2$ & 86M & 55.4G & 85.9 & 83.1\\ \midrule
Swin-S & $224^2$ & 50M & 8.7G & 436.9 & 83.0\\
Swin-B & $224^2$ & 88M & 15.4G & 278.1 & 83.5\\
Swin-B & $384^2$ & 88M & 47.0G & 84.7 & 84.5\\ \midrule
CP-ViT-B/16 (12) & $384^2$ & 88M & 57.2G & 85.1 & 79.2\\
CP-ViT-B/16 (10) & $384^2$ & 75M & 48.7G & 96.1 & 76.8\\
CP-Swin-B & $224^2$ & 89M & 16.8G & 272.3 & 84.3\\
CP-Swin-B & $384^2$ & 89M & 48.9G & 81.4 & \textbf{85.6}\\
\bottomrule
\end{tabular}
}
\end{small}
\end{center}
\vskip -0.2in
\end{table}

\section{Conclusions and Future Work}
In this paper we have shown how adaptive pooling of features for a location based on context can improve the results for a transformer model, both by reducing the number of layers needed to achieve similar accuracy and by improving accuracy of models with the same number of layers. For future work we hope to apply this technique more broadly to other domains, such as speech recognition that have multi-level contextual dependencies that span different, dynamic extents. It is our hope that adaptive pooling can benefit other such domains. In addition, a common dynamic pooling mechanism across Convnets and transformers can help to simplify hybrid architectures that adapt to context, opening up new efficient design choices.

\section*{Acknowledgements}
The authors want to thank Shih-Yu Sun, Hesam Najafi Shoushtari, Kelsey Ho and many others at Apple for helpful discussions during the course of this project. We also thank the ICML reviewers for providing useful feedback.

\bibliography{ref}

\begin{thebibliography}{41}
\providecommand{\natexlab}[1]{#1}
\providecommand{\url}[1]{\texttt{#1}}
\expandafter\ifx\csname urlstyle\endcsname\relax
  \providecommand{\doi}[1]{doi: #1}\else
  \providecommand{\doi}{doi: \begingroup \urlstyle{rm}\Url}\fi

\bibitem[Ainslie et~al.(2020)Ainslie, Onta{\~n}{\'o}n, Alberti, Cvicek, Fisher,
  Pham, Ravula, Sanghai, Wang, and Yang]{ainslie2020etc}
Ainslie, J., Onta{\~n}{\'o}n, S., Alberti, C., Cvicek, V., Fisher, Z., Pham,
  P., Ravula, A., Sanghai, S., Wang, Q., and Yang, L.
\newblock {ETC}: Encoding long and structured data in transformers.
\newblock In \emph{EMNLP}, 2020.

\bibitem[Al{-}Rfou et~al.(2019)Al{-}Rfou, Choe, Constant, Guo, and
  Jones]{RfouCCGJ19}
Al{-}Rfou, R., Choe, D., Constant, N., Guo, M., and Jones, L.
\newblock Character-level language modeling with deeper self-attention.
\newblock In \emph{AAAI}, 2019.

\bibitem[Beltagy et~al.(2020)Beltagy, Peters, and Cohan]{Beltagy2020Longformer}
Beltagy, I., Peters, M.~E., and Cohan, A.
\newblock Longformer: The long-document transformer.
\newblock \emph{arXiv:2004.05150}, 2020.

\bibitem[Child et~al.(2019)Child, Gray, Radford, and
  Sutskever]{child2019sparsetransformer}
Child, R., Gray, S., Radford, A., and Sutskever, I.
\newblock Generating long sequences with sparse transformers.
\newblock \emph{arXiv:1904.10509}, 2019.

\bibitem[Coates \& Ng(2011)Coates and Ng]{NIPS2011_6c1da886}
Coates, A. and Ng, A.
\newblock Selecting receptive fields in deep networks.
\newblock In \emph{NeurIPS}, 2011.

\bibitem[Cubuk et~al.(2020)Cubuk, Zoph, Shlens, and Le]{NEURIPS2020_d85b63ef}
Cubuk, E.~D., Zoph, B., Shlens, J., and Le, Q.
\newblock Randaugment: Practical automated data augmentation with a reduced
  search space.
\newblock In \emph{NeurIPS}, 2020.

\bibitem[Dai et~al.(2017)Dai, Qi, Xiong, Li, Zhang, Hu, and Wei]{8237351}
Dai, J., Qi, H., Xiong, Y., Li, Y., Zhang, G., Hu, H., and Wei, Y.
\newblock Deformable convolutional networks.
\newblock In \emph{ICCV}, 2017.

\bibitem[Dai et~al.(2019)Dai, Yang, Yang, Carbonell, Le, and
  Salakhutdinov]{dai-2019-transformer}
Dai, Z., Yang, Z., Yang, Y., Carbonell, J., Le, Q., and Salakhutdinov, R.
\newblock Transformer-{XL}: Attentive language models beyond a fixed-length
  context.
\newblock In \emph{ACL}, 2019.

\bibitem[Deng et~al.(2009)Deng, Dong, Socher, Li, Li, and
  Fei-Fei]{DenDon09Imagenet}
Deng, J., Dong, W., Socher, R., Li, L.-J., Li, K., and Fei-Fei, L.
\newblock Imagenet: A large-scale hierarchical image database.
\newblock In \emph{CVPR}, 2009.

\bibitem[Devlin et~al.(2019)Devlin, Chang, Lee, and
  Toutanova]{devlin-2019-bert}
Devlin, J., Chang, M.-W., Lee, K., and Toutanova, K.
\newblock {BERT}: Pre-training of deep bidirectional transformers for language
  understanding.
\newblock In \emph{NAACL-HLT}, 2019.

\bibitem[Dosovitskiy et~al.(2021)Dosovitskiy, Beyer, Kolesnikov, Weissenborn,
  Zhai, Unterthiner, Dehghani, Minderer, Heigold, Gelly, Uszkoreit, and
  Houlsby]{dosovitskiy2021an}
Dosovitskiy, A., Beyer, L., Kolesnikov, A., Weissenborn, D., Zhai, X.,
  Unterthiner, T., Dehghani, M., Minderer, M., Heigold, G., Gelly, S.,
  Uszkoreit, J., and Houlsby, N.
\newblock An image is worth 16x16 words: Transformers for image recognition at
  scale.
\newblock In \emph{ICLR}, 2021.

\bibitem[He et~al.(2014)He, Zhang, Ren, and Sun]{HeZR014}
He, K., Zhang, X., Ren, S., and Sun, J.
\newblock Spatial pyramid pooling in deep convolutional networks for visual
  recognition.
\newblock In \emph{ECCV}, 2014.

\bibitem[He et~al.(2016)He, Zhang, Ren, and Sun]{He2016DeepRL}
He, K., Zhang, X., Ren, S., and Sun, J.
\newblock Deep residual learning for image recognition.
\newblock In \emph{CVPR}, 2016.

\bibitem[Ho et~al.(2019)Ho, Kalchbrenner, Weissenborn, and
  Salimans]{ho2019axial}
Ho, J., Kalchbrenner, N., Weissenborn, D., and Salimans, T.
\newblock Axial attention in multidimensional transformers.
\newblock \emph{arXiv:1912.12180}, 2019.

\bibitem[Jia et~al.(2012)Jia, Huang, and Darrell]{6248076}
Jia, Y., Huang, C., and Darrell, T.
\newblock Beyond spatial pyramids: Receptive field learning for pooled image
  features.
\newblock In \emph{CVPR}, 2012.

\bibitem[Katharopoulos et~al.(2020)Katharopoulos, Vyas, Pappas, and
  Fleuret]{katharopoulos20}
Katharopoulos, A., Vyas, A., Pappas, N., and Fleuret, F.
\newblock Transformers are {RNNs}: Fast autoregressive transformers with linear
  attention.
\newblock In \emph{ICML}, 2020.

\bibitem[Kitaev et~al.(2020)Kitaev, Kaiser, and Levskaya]{Kitaev2020Reformer}
Kitaev, N., Kaiser, L., and Levskaya, A.
\newblock Reformer: The efficient transformer.
\newblock In \emph{ICLR}, 2020.

\bibitem[Krizhevsky(2009)]{Krizhevsky09learningmultiple}
Krizhevsky, A.
\newblock Learning multiple layers of features from tiny images.
\newblock Technical report, University of Toronto, 2009.

\bibitem[Li et~al.(2019{\natexlab{a}})Li, Jin, Xuan, Zhou, Chen, Wang, and
  Yan]{LiJXZCWY19}
Li, S., Jin, X., Xuan, Y., Zhou, X., Chen, W., Wang, Y., and Yan, X.
\newblock Enhancing the locality and breaking the memory bottleneck of
  transformer on time series forecasting.
\newblock In \emph{NeurIPS}, 2019{\natexlab{a}}.

\bibitem[Li et~al.(2019{\natexlab{b}})Li, Kaiser, Bengio, and
  Si]{pmlr-v97-li19e}
Li, Y., Kaiser, L., Bengio, S., and Si, S.
\newblock Area attention.
\newblock In \emph{ICML}, 2019{\natexlab{b}}.

\bibitem[Liu et~al.(2021)Liu, Lin, Cao, Hu, Wei, Zhang, Lin, and
  Guo]{liu2021Swin}
Liu, Z., Lin, Y., Cao, Y., Hu, H., Wei, Y., Zhang, Z., Lin, S., and Guo, B.
\newblock Swin transformer: Hierarchical vision transformer using shifted
  windows.
\newblock In \emph{ICCV}, 2021.

\bibitem[Luo et~al.(2016)Luo, Li, Urtasun, and Zemel]{NIPS2016_c8067ad1}
Luo, W., Li, Y., Urtasun, R., and Zemel, R.
\newblock Understanding the effective receptive field in deep convolutional
  neural networks.
\newblock In \emph{NeurIPS}, 2016.

\bibitem[Mahoney(2009)]{MattMahoney09}
Mahoney, M.
\newblock Large text compression benchmark.
\newblock \emph{http://mattmahoney.net/dc/textdata}, 2009.

\bibitem[Pintea et~al.(2021)Pintea, T{\"o}men, Goes, Loog, and {van
  Gemert}]{Pintea21}
Pintea, S., T{\"o}men, N., Goes, S., Loog, M., and {van Gemert}, J.
\newblock Resolution learning in deep convolutional networks using scale-space
  theory.
\newblock \emph{IEEE Transactions on Image Processing}, 30:\penalty0 8342 --
  8353, 2021.

\bibitem[Qiu et~al.(2020)Qiu, Ma, Levy, Yih, Wang, and Tang]{blockwise}
Qiu, J., Ma, H., Levy, O., Yih, W.-t., Wang, S., and Tang, J.
\newblock Blockwise self-attention for long document understanding.
\newblock In \emph{EMNLP}, 2020.

\bibitem[Rao et~al.(2021)Rao, Zhao, Liu, Lu, Zhou, and
  Hsieh]{rao2021dynamicvit}
Rao, Y., Zhao, W., Liu, B., Lu, J., Zhou, J., and Hsieh, C.-J.
\newblock {DynamicViT}: Efficient vision transformers with dynamic token
  sparsification.
\newblock In \emph{NeurIPS}, 2021.

\bibitem[Roy et~al.(2022)Roy, Saffar, Vaswani, and Grangier]{TACL2405}
Roy, A., Saffar, M., Vaswani, A., and Grangier, D.
\newblock Efficient content-based sparse attention with routing transformers.
\newblock \emph{Transactions of the Association for Computational Linguistics},
  9\penalty0 (0):\penalty0 53--68, 2022.

\bibitem[Ryoo et~al.(2021)Ryoo, Piergiovanni, Arnab, Dehghani, and
  Angelova]{ryoo2021tokenlearner}
Ryoo, M.~S., Piergiovanni, A., Arnab, A., Dehghani, M., and Angelova, A.
\newblock Tokenlearner: Adaptive space-time tokenization for videos.
\newblock In \emph{NeurIPS}, 2021.

\bibitem[Shelhamer et~al.(2019)Shelhamer, Wang, and Darrell]{Shelhamer19}
Shelhamer, E., Wang, D., and Darrell, T.
\newblock Blurring the line between structure and learning to optimize and
  adapt receptive fields.
\newblock \emph{arXiv:1904.11487}, 2019.

\bibitem[Sukhbaatar et~al.(2019)Sukhbaatar, Grave, Bojanowski, and
  Joulin]{sukhbaatar2019}
Sukhbaatar, S., Grave, E., Bojanowski, P., and Joulin, A.
\newblock Adaptive attention span in transformers.
\newblock In \emph{ACL}, 2019.

\bibitem[Tay et~al.(2020)Tay, Bahri, Yang, Metzler, and Juan]{Tay2020SparseSA}
Tay, Y., Bahri, D., Yang, L., Metzler, D., and Juan, D.-C.
\newblock Sparse sinkhorn attention.
\newblock In \emph{ICML}, 2020.

\bibitem[Tomen et~al.(2021)Tomen, Pintea, and Van~Gemert]{tomen21a}
Tomen, N., Pintea, S.-L., and Van~Gemert, J.
\newblock Deep continuous networks.
\newblock In \emph{ICML}, 2021.

\bibitem[Touvron et~al.(2021)Touvron, Cord, Douze, Massa, Sablayrolles, and
  Jegou]{touvron21a}
Touvron, H., Cord, M., Douze, M., Massa, F., Sablayrolles, A., and Jegou, H.
\newblock Training data-efficient image transformers \& distillation through
  attention.
\newblock In \emph{ICML}, 2021.

\bibitem[Vaswani et~al.(2017)Vaswani, Shazeer, Parmar, Uszkoreit, Jones, Gomez,
  Kaiser, and Polosukhin]{NIPS2017_3f5ee243}
Vaswani, A., Shazeer, N., Parmar, N., Uszkoreit, J., Jones, L., Gomez, A.~N.,
  Kaiser, L.~u., and Polosukhin, I.
\newblock Attention is all you need.
\newblock In \emph{NeurIPS}, 2017.

\bibitem[Wang et~al.(2020)Wang, Li, Khabsa, Fang, and Ma]{wang2020linformer}
Wang, S., Li, B., Khabsa, M., Fang, H., and Ma, H.
\newblock Linformer: Self-attention with linear complexity.
\newblock \emph{arXiv:2006.04768}, 2020.

\bibitem[Wang et~al.(2021)Wang, Xie, Li, Fan, Song, Liang, Lu, Luo, and
  Shao]{Wang_2021_ICCV}
Wang, W., Xie, E., Li, X., Fan, D.-P., Song, K., Liang, D., Lu, T., Luo, P.,
  and Shao, L.
\newblock Pyramid vision transformer: A versatile backbone for dense prediction
  without convolutions.
\newblock In \emph{ICCV}, 2021.

\bibitem[Wang et~al.(2018)Wang, Girshick, Gupta, and He]{8578911}
Wang, X., Girshick, R., Gupta, A., and He, K.
\newblock Non-local neural networks.
\newblock In \emph{CVPR}, 2018.

\bibitem[Yang et~al.(2018)Yang, Tu, Wong, Meng, Chao, and Zhang]{yang-2018}
Yang, B., Tu, Z., Wong, D.~F., Meng, F., Chao, L.~S., and Zhang, T.
\newblock Modeling localness for self-attention networks.
\newblock In \emph{EMNLP}, 2018.

\bibitem[Ye et~al.(2019)Ye, Guo, Gan, Qiu, and Zhang]{1911-04070}
Ye, Z., Guo, Q., Gan, Q., Qiu, X., and Zhang, Z.
\newblock Bp-transformer: Modelling long-range context via binary partitioning.
\newblock \emph{arXiv:1911.04070}, 2019.

\bibitem[Zaheer et~al.(2020)Zaheer, Guruganesh, Dubey, Ainslie, Alberti,
  Ontanon, Pham, Ravula, Wang, Yang, et~al.]{zaheer2020bigbird}
Zaheer, M., Guruganesh, G., Dubey, K.~A., Ainslie, J., Alberti, C., Ontanon,
  S., Pham, P., Ravula, A., Wang, Q., Yang, L., et~al.
\newblock Big bird: Transformers for longer sequences.
\newblock In \emph{NeurIPS}, 2020.

\bibitem[Zhang et~al.(2021)Zhang, Dai, Yang, Xiao, Yuan, Zhang, and
  Gao]{Zhang_2021_ICCV}
Zhang, P., Dai, X., Yang, J., Xiao, B., Yuan, L., Zhang, L., and Gao, J.
\newblock Multi-scale vision longformer: A new vision transformer for
  high-resolution image encoding.
\newblock In \emph{ICCV}, 2021.

\end{thebibliography}
\bibliographystyle{icml2022}

\clearpage
\appendix

\setcounter{figure}{0}
\setcounter{table}{0}

\makeatletter 
\renewcommand{\thefigure}{S\@arabic\c@figure}
\renewcommand{\thetable}{S\arabic{table}}
\makeatother

\section*{Supplementary Material}

\section{ContextPool in ConvNets}
We show our \emph{ContextPool} module can be easily applied to convolutional neural networks. A classical ConvNet is composed of alternating layers of convolution and pooling. After convolution at each layer (often followed by some activation function), assume we have a feature map $\bm{X} \in \mathbb{R}^{h \times w \times c}$ where $h,w,c$ are the height, width, and the number of channels. For a spatial location $(i,j)$ on the feature map $\bm{X}$, we use $\bm{x}_{i,j}$ to denote the corresponding feature vector at that location. The feature map $\bm{X}$ is then passed to the pooling layer, which aggregates the contextual information within a set of local regions $R$, producing a pooled feature map $\bm{Y}$ of smaller size. For the pooling function, common options include average pooling $f_{ave}()$ and max pooling $f_{max}()$. For example, we can have average pooled features $\bm{y}_k$ as:
\begin{equation}
\bm{y}_k = f_{ave}(\bm{X}|R_k) = \frac{1}{|R_k|} \sum_{(i,j) \in R_k} \bm{x}_{i,j},
\label{eqs1}
\end{equation}
where $R_k$ is the pooling region $k$ in feature map $\bm{X}$.

There are two main drawbacks with the standard average pooling function: 1) The pooling region $R_k$ is predefined (\eg,~$3 \times 3$), thus the receptive field remains fixed for each location. However, this is undesirable to encode the contexts or semantics over spatial locations because different locations may correspond to objects with varying scales. 2) The pooing function pays equal attention to all positions in a receptive field, which is usually not the case~\cite{NIPS2016_c8067ad1}. Our ContextPool method addresses these drawbacks by using \emph{learned} pooling weights and support size for each location, aiming to capture meaningful context with varying scale.

\begin{figure}[!t]
\vskip 0.2in
\begin{center}
\centerline{\includegraphics[width=1.0\columnwidth]{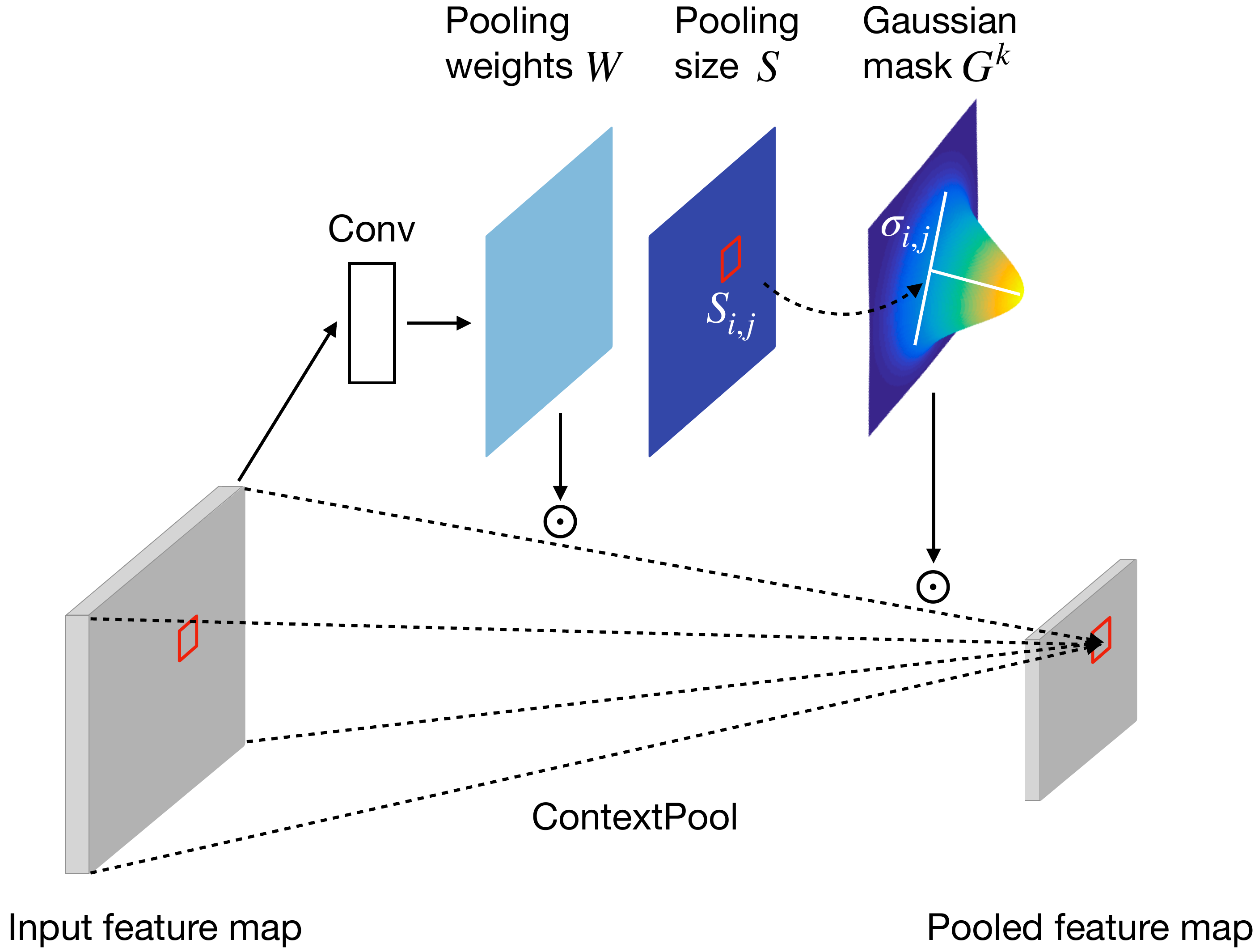}}
\caption{Illustration of our ContextPool module in ConvNets.}
\label{fig:ContextPool_CNN}
\end{center}
\vskip -0.2in
\end{figure}

Specifically, we learn the normalized maps of pooling weights $W \in \mathbb{R}^{h \times w}$ and pooling sizes $S \in \mathbb{R}^{h \times w}$ together for all positions (see Fig.~\ref{fig:ContextPool_CNN}). Both maps are conditioned on the input feature map $\bm{X}$,~\ie,~$\{ W,S \}=m(\bm{X})$ with the same spatial resolution with $\bm{X}$. Note the pooling weights $W$ are normalized by a softmax function in order to apply effective weighting over different positions during pooling. While we learn normalized pooling size $S_{i,j} \in [0,1]$ mainly to make its learning invariant to feature map size. This way, during the actual pooling for position $(i,j)$, we can easily transform $S_{i,j}$ to the standard deviation $\sigma_{i,j}=r \cdot S_{i,j} \cdot (w+h)/2$ of a Gaussian mask $G \sim \mathcal{N}(i,j,\sigma_{i,j}^2,\sigma_{i,j}^2)$. Here $r$ is an empirically set scalar (say 0.05), and $G \in \mathbb{R}^{h \times w}$ imposes spatial locality for pooling.

Finally, given the pooling weights $W$ and Gaussian mask $G^k$ for the pooling center $k$, our ContextPool module aggregates information across all the spatial positions in input feature map $\bm{X}$. In other words, ContextPool operates on the 2D spatial domain for the 3D input $\bm{X}$, and the operation remains the same across the channel dimension:
\begin{equation}
\bm{y}_k = f_{ave}(\bm{X} \odot \gamma(W) \odot \gamma(G^k)) = \sum_{i,j} \bm{x}_{i,j} \cdot W_{i,j} \cdot G^k_{i,j},
\label{eqs2}
\end{equation}
where $\gamma(\cdot)$ is a broadcasting function to accommodate element-wise multiplication $\odot$. The normalization factor is set as $C(\bm{X}) = \sum_{i,j} W_{i,j} \cdot G^k_{i,j}$.

In practice, the prediction function $m(\cdot)$ for pooling weights $W$ and sizes $S$ is implemented by applying two convolutional layers over the feature map $\bm{X}$. During training, the convolutional kernels for both the main network and ContextPool are learned simultaneously. We show our ContextPool is pretty lightweight with small increase in model size, and is able to consistently improve performance. We validate this on two common benchmarks for image classification, as we now demonstrate.

\section{Results on Image Classification}
\textbf{CIFAR-10 dataset}
We first evaluate ConvNets equipped with ContextPool (CP) for image classification on CIFAR-10 dataset~\cite{Krizhevsky09learningmultiple}. CIFAR-10 consists of 60k images with 10 classes. We follow the standard training and testing protocol, using 50k images for training a ResNet~\cite{He2016DeepRL} and 10k images for testing. 

Table~\ref{supp_tb1} shows the ResNet-44 baseline with regular pooling function obtains 92.9\% accuracy on CIFAR-10. While DCN and N-Jet-based methods are parameter-efficient when learning adaptive kernel size using Gaussian derivative filters. They show success of learning data-dependent receptive fields, but the performances are not as competitive as those of other methods. Note the results are from the original papers using only small model sizes. It remains unclear how performance scales with increasing model size.
On the other hand, deformable ConvNets~\cite{8237351} learn spatial offsets for the sampling locations of convolution and pooling operations, offering an alternative way for learning adaptive receptive field. We observe that both the deformable convolution and deformable pooling modules contribute to compelling results.

In comparison, our CP-improved ResNets achieve a better trade-off between performance and parameter efficiency than deformable ConvNets. When applied to the same ResNet-44 backbone, our CP already achieves a competitive accuracy of 93.4\% at low overhead. We can further improve accuracy to 93.7\% by training a deeper network with CP. Note the resulting CP-ResNet-46 outperforms deformable ConvNets with a similar model size.

Lastly, we offer two more variants of ContextPool in the ConvNet framework. For the first variant, we only learn adaptive pooling size, with uniform pooling weights (\ie,~average pooling). This baseline is analogous to those learning methods for pooling region or receptive field~\cite{NIPS2011_6c1da886}. Another related method is spatial pyramid pooling~\cite{HeZR014}. But this method is not directly comparable because it is mainly designed to deal with input images of varying size. Table~\ref{supp_tb1} (bottom cell) shows that our pooling size learning performs slightly worse than deformable pooling~\cite{8237351}. More importantly, it is inferior to our full method due to the lack of dynamic pooling weights. When we replace our learned pooling weights with those defined by the feature similarity (as done for transformers in main paper), we see marginal improvements which indicates the need of pooling weights learning.

\begin{table}[t]
\caption{Model size and performance (\%) on CIFAR-10. Results are reported over three runs per setting.}
\label{supp_tb1}
\begin{center}
\begin{small}
\resizebox{\linewidth}{!}{
\begin{tabular}{ccc}
\toprule
Method & Size & Accuracy \\
\midrule
ResNet-44~\cite{He2016DeepRL} & 0.66M & 92.9\\
DCN~\cite{tomen21a}   & 0.47M & 89.7$\pm$0.3 \\
N-Jet-ResNet-32~\cite{Pintea21}    & 0.52M & 92.3$\pm$0.3\\ 
Deform ResNet-44 (Pool)~\cite{8237351}   & 0.68M & 93.2$\pm$0.4\\
Deform ResNet-44 (Pool+Conv)~\cite{8237351}   & 0.69M & 93.5$\pm$0.2 \\ \midrule
CP-ResNet-44  & 0.68M& 93.4$\pm$0.3\\
CP-ResNet-46   & 0.70M& \textbf{93.7$\pm$0.2}\\ \midrule
CP-ResNet-44 (learn pooling size only)  & 0.67M& 93.1$\pm$0.2\\
CP-ResNet-44 (pooling weights by fea similarity)   & 0.67M& 93.2$\pm$0.2\\
\bottomrule
\end{tabular}
}
\end{small}
\end{center}
\vskip -0.1in
\end{table}

\begin{table}[t]
\caption{Classification accuracy (\%) and model size on ImageNet.}
\label{supp_tb2}
\begin{center}
\begin{small}
\resizebox{\linewidth}{!}{
\begin{tabular}{ccccc}
\toprule
Backbone & Method & Top-1 & Top-5 & Size \\
\midrule
\multirow{3}{*}{ResNet-50} & baseline & 76.5 & 93.1 & 26.6M\\
 & Deform~\cite{8237351} & 76.6 & 93.2 & 26.8M\\
 & CP-baseline & \textbf{77.3} & \textbf{93.6} & 26.8M\\ \midrule
\multirow{3}{*}{ResNet-101} & baseline & 78.4 & 94.2 & 45.5M\\
 & Deform~\cite{8237351} & 78.4 & 94.2 & 45.8M\\
 & CP-baseline & \textbf{78.9} & \textbf{94.4} & 45.8M\\
\bottomrule
\end{tabular}
}
\end{small}
\end{center}
\vskip -0.1in
\end{table}

\textbf{ImageNet-1K dataset}
We further compare our CP-improved ResNets with the strong baseline of deformable ConvNets~\cite{8237351} on ImageNet-1K dataset. For a fair comparison, we use the same training and inference settings as in~\cite{8237351}. Table~\ref{supp_tb2} illustrates the validation-set results based on two ResNet backbones. It can be observed that our CP-ResNets achieve consistent improvements over both the baseline and deformable ConvNets, without large increase in model size. Our hypothesis is that CP benefits more from its strong context modeling capability on high-resolution ImageNet images. For future work, it would be interesting to test our approach on various image resolutions or on more types of tasks that have different needs for a context model.

\end{document}